\documentclass[10pt,twocolumn,letterpaper]{article}
\usepackage{wacv}
\usepackage{times}
\usepackage{epsfig}
\usepackage{graphicx}
\usepackage{amsmath}
\usepackage{amssymb}
\usepackage{multirow}
\usepackage{rotating}

\wacvfinalcopy 

\ifwacvfinal
\def\assignedStartPage{9876} 
\fi

\ifwacvfinal
\usepackage[breaklinks=true,bookmarks=false]{hyperref}
\else
\usepackage[pagebackref=true,breaklinks=true,colorlinks,bookmarks=false]{hyperref}
\fi

\ifwacvfinal
\else
\fi

\begin{document}

\title{Controlled Caption Generation for Images Through \\ Adversarial Attacks}

\author{Nayyer Aafaq\\
University of Western Australia\\
35 Stirling Highway, 6009, WA\\
{\tt\small nayyer.aafaq@research.uwa.edu.au}
\and
Naveed Akhtar\\
University of Western Australia\\
35 Stirling Highway, 6009, WA\\
{\tt\small naveed.akhtar@uwa.edu.au}
\and
Wei Liu\\
University of Western Australia\\
35 Stirling Highway, 6009, WA\\
{\tt\small wei.liu@uwa.edu.au}
\and
Mubarak Shah\\
University of Central Florida\\
4000 Central Florida Blvd, Orlando, Florida, USA\\
{\tt\small shah@crcv.ucf.edu}
\and
Ajmal Mian\\
University of Western Australia\\
35 Stirling Highway, 6009, WA\\
{\tt\small ajmal.mian@uwa.edu.au}
}

\maketitle

\begin{abstract}
   Deep learning is found to be vulnerable to adversarial examples. However, its adversarial susceptibility in image caption generation is under-explored. We study adversarial examples for vision and language models, which typically adopt an encoder-decoder framework consisting of two major components: a Convolutional Neural Network (\ie,~CNN) for image feature extraction and a Recurrent Neural Network (RNN) for caption generation. In particular, we investigate attacks on the visual encoder's hidden layer that is fed to the subsequent recurrent network. The existing methods either attack the classification layer of the visual encoder or they back-propagate the gradients from the language model. In contrast, we propose a GAN-based algorithm for crafting adversarial examples for neural image captioning that mimics the internal representation of the CNN such that the resulting deep features of the input image enable a controlled incorrect caption generation through the recurrent network. 
Our contribution provides new insights for understanding adversarial attacks on vision systems with language component. The proposed method employs two strategies for a comprehensive evaluation. The first examines if a neural image captioning system can be misled to output targeted image captions. The second analyzes the possibility of keywords into the predicted captions. Experiments show that our algorithm can craft effective adversarial images based on the CNN hidden layers to fool captioning framework. Moreover, we discover the proposed attack to be highly transferable. Our work leads to new robustness implications for neural image captioning. 
\end{abstract}

\vspace{-10mm}
\section{Introduction}
Deep learning,  which promises to solve highly complex problems, has achieved great success for various tasks, such as image classification, speech recognition, and machine translation. However, recent studies have unearthed a multitude of  adversarial attacks~\cite{akhtar2018threat} for deep learning models, which may hinder the adoption of this technique in security-sensitive applications~\cite{szegedy2015going, papernot2016transferability, liu2016delving, goodfellow2014explaining}. For adversarial attacks, existing works mainly consider the task of image classification, and demonstrate that it is nearly always possible to fool deep classifiers to predict incorrect label for the adversarial examples~\cite{szegedy2013intriguing}. Although the literature has also seen many techniques to defend against such attacks~\cite{tramer2017ensemble, papernot2016distillation, metzen2017detecting, meng2017magnet, madry2017towards, goodfellow2014explaining},  many of the defenses are later found to be easily broken with counter-attacks~\cite{carlini2017adversarial, he2017adversarial, carlini2016defensive}. 

\begin{figure}[t]
  \centering
     \includegraphics[width=0.9\columnwidth]{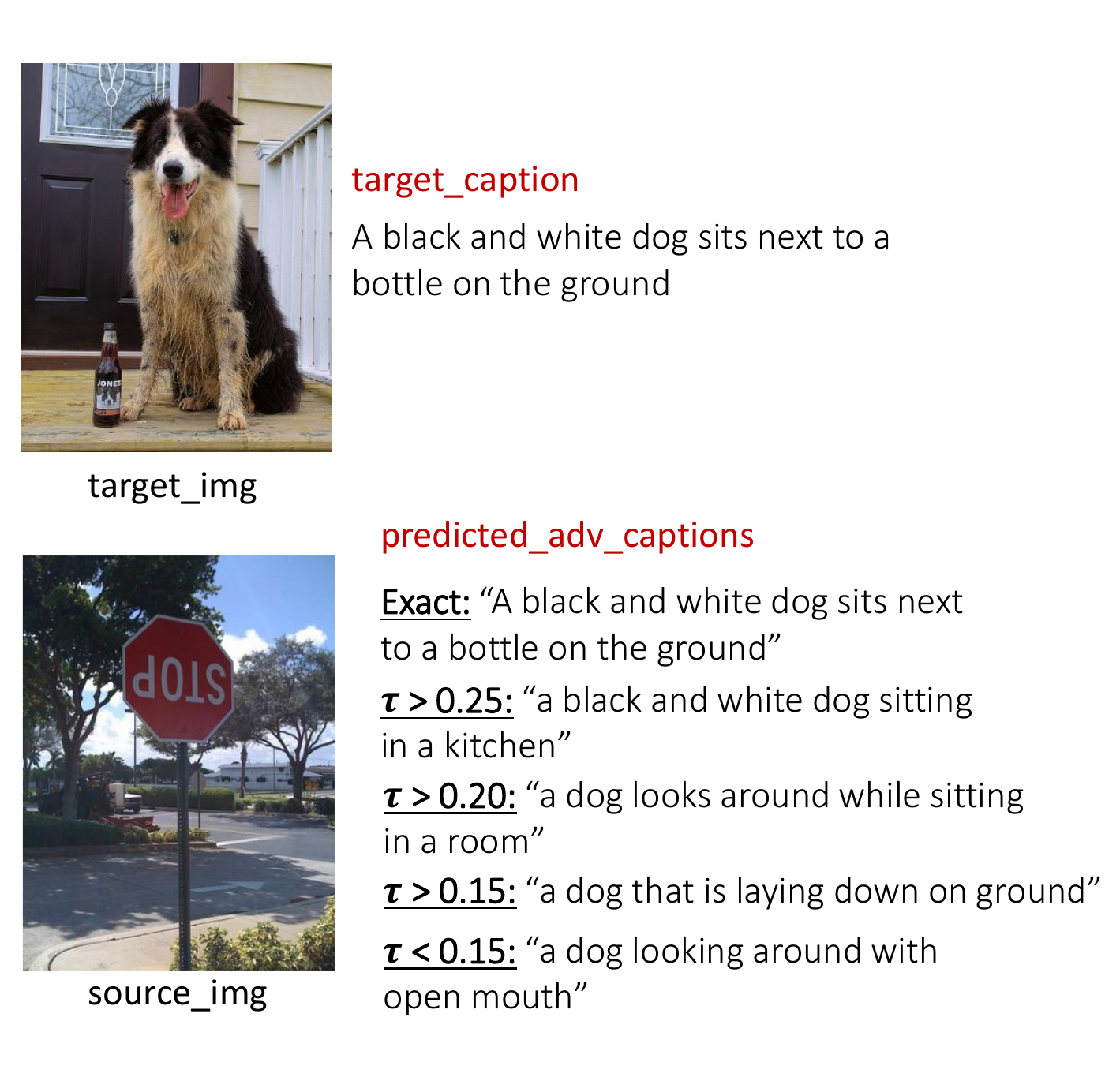}
     \caption{Examples of adversarial attacks on the Show-and-Tell~\cite{vinyals2015show} model with target image (top) and source image (bottom). We show  example results for Exact and all $\tau$ thresholds. In our experiments, we categorise all the results under $\tau < 0.15$ as failed attempts. However, as evident from the predictions, despite $\tau$ below the minimum threshold value, the predicted captions are still relevant to the target image.}
      \label{fig:introfig}
      \vspace{-5mm}
\end{figure}

More recently, there has been an increasing interest in the research community to explore whether adversarial examples are also effective against more complex vision systems~\cite{lu2017knowing, lu2017standard, lu2017no}.
For instance, among the latest results along this debate, Lu et al. showed that adversarial examples previously constructed to fool CNN-based classifiers cannot fool state-of-the-art detectors~\cite{lu2017standard}. In this work, we are interested in exploring  whether language context in language-and-vision systems offer more resistance to adversarial examples.  Our investigation incorporates the more complex systems that have both visual and language  components. 
In specific, we investigate the image captioning framework with these components. Such a framework offers two unusual  challenges for computing effective adversarial examples. First, it is not straightforward to extend the notion of adversarial examples used for a classification problem to the image captioning problem.
Classifiers map an image to a discrete label. Altering that label requires a simple adversarial objective to disrupt this mapping with another discrete label. 
If we treat the caption of an image as a discrete label, the search space of the mapping becomes overwhelmingly large, as the same information can be expressed in a number of different captions, while still being correct in all the cases. This differentiates the adversarial objective for  captioning significantly from the conventional classification attacks. 
Second, the conventional attacks on CNNs are aimed at fooling a single model. However, captioning is a multi-model problem that involves CNN-RNN architecture. Whereas the attacks on CNN models have been well studied, the attacks on RNNs are still under-explored, leaving alone the specific challenges of a `multi-model' problem in captioning. 

In the closely related literature, the first attempt to attack a captioning framework was made by Xu~\etal~\cite{xu2018fooling}. Their approach treats the output caption as a single label, thereby focusing on fooling the RNN part of the framework while keeping the CNN embeddings unchanged. The method of `Show and Fool'~\cite{chen2017attacking} introduced the notions of  `targeted caption' and `targeted keyword' in the context of adversarial attacks on the captioning framework. Recently,~\cite{xu2019exact} proposed a more restricted targeted partial captions attack to enforce targeted keywords to occur at a specific location in the caption. All these methods must know the optimization objective of the language model to generate the adversarial examples. While useful, this scheme requires the complete knowledge of the captioning framework to fool it. Moreover, it requires computing the adversarial samples anew if the visual model in the framework is replaced. Considering that captioning methods frequently rely on a variety of off-the-shelf visual models, this is undesirable.  

We propose to perturb the image based on the internal representation of the visual model used in the captioning framework. This allows us to effectively control the image caption without requiring the knowledge of the language model, which is more practical (see Fig.~\ref{fig:introfig}). 
Our attack alters the image such that the internal representation of a visual classifier gets drastically changed for the image with minimal image-space perturbations. 
Since the language model depends strongly on the internal representation of the visual model in captioning frameworks, this allows us to control the eventual caption without any knowledge of the language model. 
To generate the adversarial images we propose a GAN-based method that aims at emulating the target representation of the desired internal layer for the image. The generator is trained to compute the perturbation that results in pronounced image features of the target (incorrect) class and suppressed features of the source (correct) class.  
The source and target classes can be pre-selected in our attack. 
We summarise the main contributions of our work as follows.



\vspace{-3mm}
\begin{itemize}
\item We propose a GAN-based attack on image captioning framework that is able to alter the internal representation of the image to control the captions generated by the subsequent language model. We are the first to study the adversarial attack on image captioning system via visual encoder only.  
\vspace{-2mm}
\item We demonstrate that the captioning frameworks can be fooled with much higher success rates when attacked on the internal layers of the visual models instead of the  classification layer.
\vspace{-2mm}
\item Extensive experiments show that state-of-the-art image captioning models can be fooled by the proposed method with high success rate.
\vspace{-2mm}
\item Lastly, we explore the workings of internal mechanism of the captioning framework via  adversarial attack. 
\end{itemize}
\section{Related Work}
Below we first review work on visual captioning (\ie,~image and video captioning) followed by adversarial attacks on the  visual tasks. Lastly, we discuss the existing line of work on adversarial attacks on captioning framework.  

\subsection{Visual Captioning}
Visual captioning is a multi-modal problem which has recently gained a surge of interest, primarily due to the advancement in deep learning~\cite{lecun2015deep}, availability of large scale datasets~\cite{aafaq2019video,lin2014microsoft,xu2016msr,krishna2017dense,krishna2017visual} and computational capability enhancement of  machines. Visual captioning includes image captioning~\cite{Pan_2020_CVPR, Feng_2019_CVPR, Yang_2019_CVPR}, video captioning~\cite{zhang2020video, aafaq2019spatio, 9027096,park2020identity}, and {\em dense} image/video captioning~\cite{johnson2016densecap,krishna2017dense}. Most visual captioning approaches adopt encoder-decoder architecture while others have extended it to perform attention~\cite{xu2015show,yao2015describing,yan2019stat} over visual cues, semantic concepts or attributes~\cite{gan2017semantic,Pan_2017_CVPR}. The encoder extracts features from the input image/video. For that matter pre-trained deep neural models (\ie,~CNNs) are used as encoders in the captioning framework\cite{aafaq2021empirical}. Subsequently, recurrent networks~~\cite{cho2014learning,hochreiter1997long} or transformers~\cite{vaswani2017attention} are employed to decode the visual features into natural language sentences.

\subsection{Adversarial Examples}
Where deep learning has proposed solutions to several complex  problems, it has been proven that it is vulnerable to adversarial examples~\cite{szegedy2013intriguing}. Depending on the available information about the target model, contemporary works can be classified into three main categories \ie,~{\em white-box}, {\em gray-box} and {\em black-box} attacks. We refer the readers to \cite{akhtar2018threat} for more details.
Deep networks have performed really well on visual tasks \eg,~image classification~\cite{krizhevsky2012imagenet}, object detection and semantic segmentation~\cite{girshick2014rich,long2015fully}, sometimes even surpassing  human performance. Primarily due to this fact, most existing works on adversarial attacks mainly study the robustness of deep models for these tasks. For instance, for image classification, many techniques have been proposed for computing adversarial examples; \eg~Fast Gradient Sign Method (FGSM)~\cite{goodfellow2014explaining}, Iterative Fast Gradient Sign Method (I-FGSM)~\cite{kurakin2016adversarial}, optimisation based methods Carlini and Wager attack~\cite{carlini2017towards}, box-constrained L-BFGS~\cite{szegedy2013intriguing}, DeepFool~\cite{moosavi2016deepfool} and others. These methods demonstrate that CNN-based deep visual classifiers can be easily fooled with adversarial examples. Similarly, other CNN-based visual model for \eg object detection and semantic segmentation have also been demonstrated to be susceptible to adversarial attacks~\cite{xie2017adversarial,lu2017no,song2018physical,arnab2018robustness}. 

It is easy to see that all the aforementioned methods share the same trait that they are at CNN models for classification, which are differentiable programs. Hence, their loss function gradients with respect to input image can easily be computed with backpropagation to generate  adversarial examples. 
Different from CNNs, another line of works explore the adversarial examples for recurrent neural networks for text processing~\cite{papernot2016crafting,jia2017adversarial} and deep reinforcement learning~\cite{lin2017tactics,kos2017delving,huang2017adversarial}. There, the backpropagation of gradients and model's fooling objective become more complicated due to the underlying architecture and objectives of the original models. In our work, we study the adversarial examples for image captioning framework~\ie,~vision-language model that faces the challenges fooling both CNNs and RNNs.  

\begin{figure*}[t]
  \centering
     \includegraphics[width=0.8\textwidth]{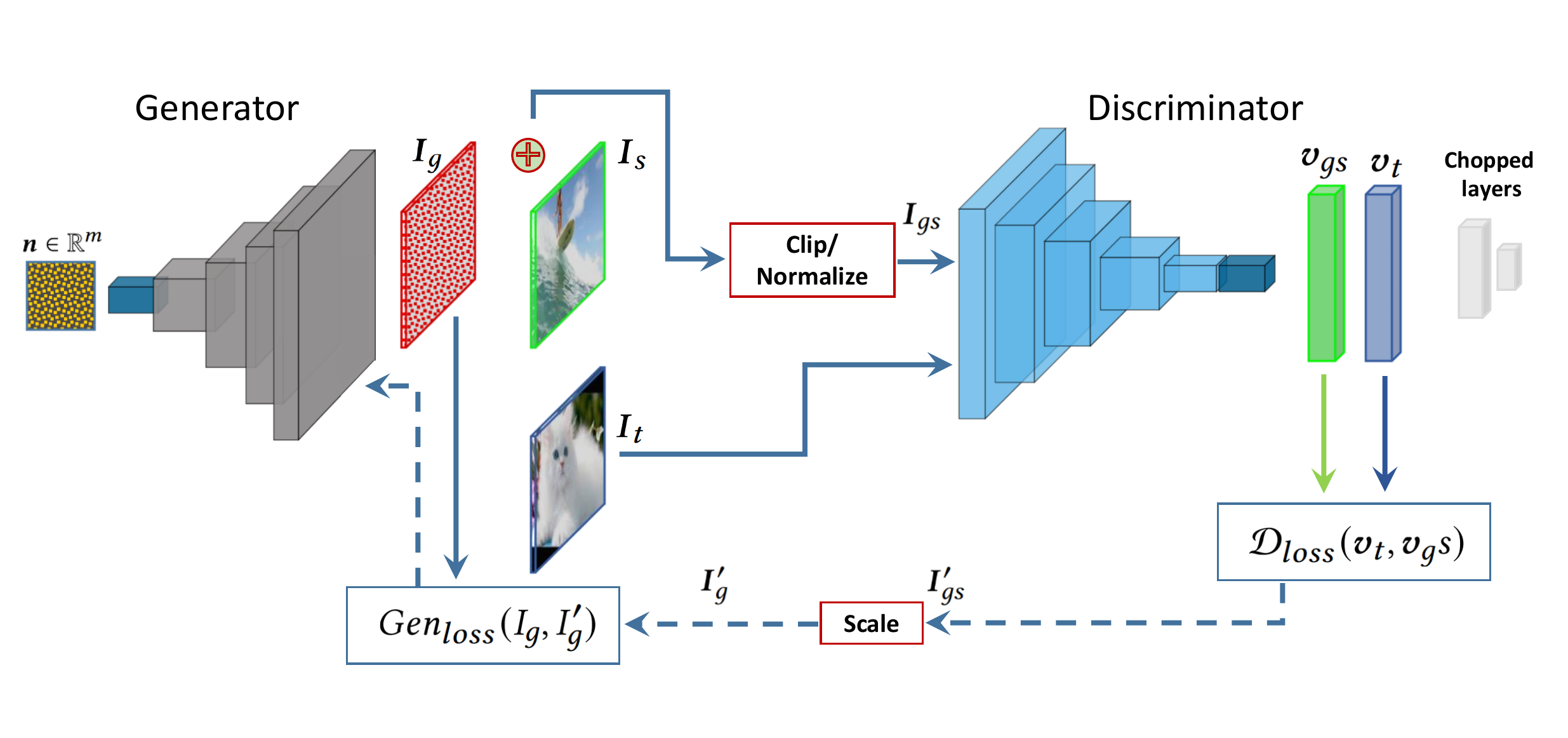}
     \vspace{-4mm}
     \caption{{\em Proposed Methodology}: Overall architecture for generating adversarial examples for image captioning framework is shown. A fixed pattern, sampled from a uniform distribution \ie,~$\boldsymbol n \in \mathbb R^m$ is passed through the generator. The output of the generator is added to the source image \ie,~$\boldsymbol I_s$ and the scaled result is passed through the discriminator to get the deep representation $\boldsymbol v_{gs}$ of the desired $j^{\text{th}}$ layer.  Similarly, target image \ie,~$\boldsymbol I_t$ is passed through the discriminator to get $\boldsymbol v_t$. Discriminator computes and back-propagates the gradients to update the input image $\boldsymbol I_{gs}^\prime$. After re-scaling and subtracting the original image, updated perturbations $\boldsymbol I_g^\prime$ are separated and generator gradients are updated. The output of the trained generator is then added to the source image to generate the captions  similar to the target image descriptions.}
      \label{fig:schema}
      \vspace{-3mm}
\end{figure*}

\subsection{Adversarial Examples For Captioning Tasks}
Most of the contemporary captioning frameworks compose of CNN followed by RNN. The pre-trained CNNs are used to extract features from the visual input. The extracted features from a selected hidden layer of the CNN are then fed to the sequence model for natural language caption for the  visual input. In contrast to CNNs, the captioning framework (\ie,~CNN+RNN) carries additional temporal dependency in the generated output \ie,~words. Due to the temporal relationship among the generated words, it is much more challenging to compute the desired (adversarial) gradients for the overall framework.  
Xu \etal~\cite{xu2018fooling}, are the first  to attempt an attack on captioning framework. In their approach, they treated the output sentences as single label, thereby   altering a complete sentences. Moreover, they focused on generating adversarial examples to fool the RNN part of the model, while retaining the original CNN embeddings. The method `Show and Fool'~\cite{chen2017attacking} proposed `targeted captions' and `targeted keywords' attacks on the captioning framework. The attack of targeted captions generate complete captions for the input image. In its `targeted keywords' attack, the predicted sentence should include the targeted keywords. Recently,~\cite{xu2019exact} proposed more restricted targeted partial captions attack that can enforce the targeted keywords to occur at specific locations. These methods leverage from language model objective functions to generate the adversarial examples. Though successful, this scheme requires complete model information for the whole framework for successful fooling. This is often not available.


\section{Methodology}
\subsection{Problem Formulation}
Before articulating our methodology, we will first briefly touch the generic strategy of adversarial attack for classification models, explaining how is this different from the adversarial attack on a captioning framework. 
Let $\boldsymbol I \in \mathbb R^m$ be a sample of a distribution $\mathcal I$ over the natural images and $\phi_{\theta}(\boldsymbol I)$ be a deep visual classification model
that maps $\boldsymbol I$ to its correct label $y$. The common aim of
generating perturbations in adversarial settings is to compute $\boldsymbol p \in \mathbb R^m$ that satisfies the constraint
$$\phi_{\theta}(\boldsymbol I + \boldsymbol p) = \hat{y} \hspace{2mm} s.t. \hspace{2mm} \hat{y} \neq y,~ ||\boldsymbol p||_p \leq \eta ,$$
where $y$ and $\hat{y}$ denote the true and incorrect labels respectively, $||.||_p$ denotes the $\ell_p$ norm that is  constrained by $\eta$ in the optimization problem. In the  above equation, restricting $\hat{y}$ to a pre-defined label results in a targeted adversarial attack.

For a given CNN+RNN captioning framework, the targeted caption is denoted by 
$$S = (S_1, S_2, ..., S_N),$$
where $S$ indicates the words in vocabulary $\mathcal V$ with $S_1$ and $S_N$ representing start and end of caption symbols and $N$ is the length of the caption. Note that caption length \ie,~$N$ is variable but it does not exceed pre-defined maximum length of the caption. The posterior probability of the caption is computed as
$$P(\boldsymbol S|\boldsymbol I + \boldsymbol p, \boldsymbol \theta) = \prod_{t=1}^N P(\boldsymbol S_t|\boldsymbol S_{<t}, \boldsymbol I + \boldsymbol p; \boldsymbol \theta),$$
where $\boldsymbol I$ represents the benign image and $\boldsymbol p$ shows the perturbations. In order to predict the targeted caption $S$, captioning system needs to maximize the probability of the caption $S$ conditioned on the image $\boldsymbol I + \boldsymbol p$ among all possible captions set $\Psi$, i.e.,
$$log \hspace{1mm} P(S|\boldsymbol I + \boldsymbol p) = \max_{{S_\circ} \in \Psi} log \hspace{1mm} P(S_\circ| \boldsymbol I + \boldsymbol p).$$

In captioning frameworks (\ie,~CNN + RNN), the probability is usually computed by RNN cell $\varphi$, with its hidden state $h_{t-1}$ and input word $S_{t-1}$
$$z_t = \varphi(h_{t-1}, S_{t-1}),$$
where $z_t$ represents vector of logits for every possible word in the vocabulary $\mathcal{V}$. Most existing methods try to maximize the captions probabilities by directly using its negative log probabilities as a loss function where as the RNN cell hidden state at $t^{\text{th}}$ time step is initialised with $h_{t-1}$ from the CNN model. For CNN model here, an existing pre-trained model \eg,~ResNet-152 is employed. In the captioning models, the CNN is not trained but for input image one may compute perturbations by backpropagating error from the LSTM to the CNN input. The operation needs to be repeated for all time steps to allow the LSTM to update weights using backpropagation through time (BPTT) across a sequence of the internal vector representations of input images. Another possibility is to fool the captioning framework  by employing classification-base attack on the CNN model and then feed the perturbed image hidden state to the LSTM and evaluate the robustness of the captioning framework. However, as LSTM uses hidden state from the deep CNN, and not the classification layer logits, this simple strategy can not be expected to be  effective.

Intuitively, there is a single pre-trained CNN model and a sequence of LSTM models, one for each time step. We want to apply the perturbations to the CNN model and pass on the output of each input image to the LSTM as a single time step. For that matter, rather than manipulating the class label of the image, we manipulate the internal representation of the CNN $\phi_\theta(.)$ to conduct fooling of the captioning framework. 
This allows us to compute the perturbations independently of the RNN stage of the framework. 

Let $\phi_j(\boldsymbol I)$ be the activations of the $j^{\text{th}}$ layer of the classification network $\phi$ when processing the image $\boldsymbol I$.  We formulate the feature reconstruction loss as the squared and normalized Euclidean distance between the feature representations:
$$\mathcal L_\phi^j (\hat{y}, y) = ||\phi_j(\hat{y}) - \phi_j(y)||^2_2.$$

As demonstrated in \cite{mahendran2015understanding},  finding an image $\hat{y}$ that minimizes the feature reconstruction loss for early layers tends to produce images that are visually indistinguishable from $y$. As we reconstruct the image from higher layers, image content and overall spatial structure are preserved. 
Using a feature reconstruction loss for training our generator network encourages the source image $\boldsymbol I_s$ to produce similar internal representations as of the target image $\boldsymbol I_t$.

\subsection{Computing Generative Perturbations}
To compute our adversarial perturbations, we sample a fixed pattern $\mathcal Z \in [0,1]^n$  from a uniform distribution $U[0,1]^n$. This sample is fed to the generator $G_\theta$ to compute the perturbation. The output of the generator $G_\theta(\mathcal Z)$ is first scaled to have a fixed norm. Henceforth, we denote the generator output as $I_g$ for brevity. The scaled output is then added to the source image, producing $I_{gs}$ with an aim to have the internal layer features of the image same as the target image $I_t$, see Fig.~\ref{fig:schema}. This means, that the language model in turn will generate the same captions as for the target image despite the fact that the two images are altogether different. In this way, we can control the output of the language model without any information about the model.  

The discriminator forms the second part of our overall GAN setup. Before feeding both  images to the discriminator, we clip them to keep them in the valid dynamic range of the  images on which the discriminator is trained. We feed the clipped images $I_{gs}$ and $I_t$ to the discriminator network $\mathcal D$ to obtain the output feature vectors \ie,~$\boldsymbol v_{gs}$ and $\boldsymbol v_t$ of the $j^{\text{th}}$ layer of the discriminator. We employ pre-trained CNN as the discriminator in our approach. We chop off the classification layer (all undesired layers from the top) to obtain the desired \ie,~$j^{\text{th}}$ layer features as output such that; 

$$\boldsymbol n \in \mathbb R^{100},$$
$$ I_g = G(\boldsymbol n); \hspace{2mm} \boldsymbol I_g \in \mathbb R^{W \times H \times C} \hspace{2mm} \rightarrow [-1, 1],$$
$$\boldsymbol I_s \in \mathbb R^{W \times H \times C} \hspace{2mm} \rightarrow [-1, 1],$$
$$ \boldsymbol v_s = \mathcal{D}(clip(\boldsymbol I_g * \epsilon + \boldsymbol I_s)_{-1}^{1}) \in \mathbb R^{m},$$
$$\boldsymbol v_t = \mathcal{D}(\boldsymbol I_t)  \in \mathbb R^{m},$$
$$\boldsymbol I_g' = \boldsymbol I_s' - \boldsymbol I_s,$$
where $\epsilon$ is a pre-fixed scaling factor and $clip(.)_{-1}^{1}$ performs clipping in the range $[-1,1]$.

\subsection{Loss Functions}
Our discriminator is able to compute the loss of the two vectors from $j^{\text{th}}$ layer as
\vspace{-1mm}
$$\mathcal D_{loss}(\boldsymbol v_t, \boldsymbol v_{gs}) = \sum_{k=1}^K ||\phi_j(\boldsymbol v_t) - \phi_j(\boldsymbol v_{gs})||^2_2.$$
\vspace{-1mm}
We back propagate the gradients from the loss to update  $I_{gs}$. Subsequently, 
$I_g^\prime$ is taken out to further get the generator loss as
\vspace{-1mm}
$$Gen_{loss} (I_g, I_g') = - \sum_{k=1}^K {I_g}_k log({I_g'}_k).$$
 \vspace{-1mm}
\subsection{Implementation Details} 
We use ADAM optimizer~\cite{kingma2015adam} to minimize our generator and discriminator losses. We set the learning rate to $1 \times 10^{-4}$. The generator is trained for 300 epochs. All our experiments are performed on NVIDIA TITAN RTX GPU. For each class experiments, we train the generator on train set and validate it by using all the images of the source class in the validation set of $5000$ images. Details of the dataset is provided below. We perform testing of our algorithm under gray-box settings where the attack is launched on the captioning framework by manipulating the visual input only. To that end, the baseline is achieved by adopting a vanilla CNN and LSTM combination. 
We also test our technique for the popular Show-and-Tell~\cite{vinyals2015show}, and Show-Attend-and-Tell~\cite{xu2015show} image captioning models. 

\section{Experiments and Results}
In this section, we discuss the attack performance of our method on two popular captioning frameworks \ie,~Show-and-Tell~\cite{vinyals2015show}, and Show-Attend-and-Tell~\cite{xu2015show}. We also perform experiments to explore the internal mechanism of the captioning framework for better understanding of the captioning models. 

\subsection{Dataset}
We evaluate our technique on the benchmark dataset for image captioning \ie,~Microsoft COCO 2014 (MS COCO)~\cite{lin2014microsoft}. We split the images into $113287$, $5000$ and $5000$ for training, validation and test set following~\cite{karpathy2015deep}. 
For  training purposes, we further split the images based on their classes
and availability of their sufficient  number of examples in the training  set for generator training.

\subsection{Metrics}
We employ the popular BLEU-1, BLEU-2, BLEU-3, BLEU-4, CIDEr, METEOR, ROUGE and SPICE metrics to evaluate the predicted captions for the targeted adversarial captions. Except CIDEr witch is specifically designed for image captioning, the literature lacks metrics to evaluate captioning system. Therefore these metrics, that are widely used by NLP community, have been adopted by image captioning systems for quality measurement of the automated generated captions. 
\begin{table*}[htbp]
  \small
  \centering
  \caption{Results of adversarial attacks on the Show-and-Tell~\cite{vinyals2015show} model with  $\epsilon$ = [0.05, 0.1, 0.2, 0.25]. $\tau$ indicates threshold for METEOR metric score for similarity comparison between predicted and targeted caption. Accuracy shows the \%age of examples successfully attacked. Average of $||\delta||_2$ is taken over all the successful examples.}
    \begin{tabular}{c|c|c|c|c|c|c|c|c|c}
    \hline
    \multirow{2}{*}{Metric} & \multirow{1.5}{*}{Threshold } & \multicolumn{2}{c|}{$\epsilon = 0.05$} & \multicolumn{2}{c|}{$\epsilon = 0.1$} & \multicolumn{2}{c|}{$\epsilon = 0.2$} & \multicolumn{2}{c}{$\epsilon = 0.3$} \\
 & ($\tau$) & Accuracy & Avg. $||\delta||_2$ & Accuracy & Avg. $||\delta||_2$ & Accuracy & Avg. $||\delta||_2$ & Accuracy & Avg. $||\delta||_2$ \\
    \hline
    \multirow{3}{*}{METEOR} & 0.15 & 71.3 & 1.6251 & 79.2 & 2.6874 & 84.6 & 4.5894 & 90.1 & 6.2670 \\
    & 0.20 & 64.1 & 1.7399 & 68.7 & 2.6984 & 75.3 & 4.8625 & 81.9 & 6.5719 \\
    & 0.25 & 50.4 & 1.8456 & 62.4 & 2.7311 & 69.7 & 4.9347 & 73.5 & 6.7427 \\
    \hline
    Exact Match & 1.0 & 31.9 & 1.7621 & 47.7 & 2.9523 & 55.6 & 4.8425 & 64.1 & 6.5554 \\
    \hline
    \end{tabular}%
  \label{tab:SAAT}%
\end{table*}%

\begin{table*}[htbp]
    \small
    \centering
  \caption{Results of adversarial attacks on the Show-Attend-and-Tell~\cite{xu2015show} model with  $\epsilon$ = [0.05, 0.1, 0.2, 0.25]. $\tau$ indicates threshold for METEOR metric score for similarity comparison between predicted and targeted caption. Accuracy shows the \%age of examples successfully attacked. Average of $||\delta||_2$ is taken over all the successful examples.}
    \begin{tabular}{c|c|c|c|c|c|c|c|c|c}
    \hline
    \multirow{2}{*}{Metric} & \multirow{1.5}{*}{Threshold } & \multicolumn{2}{c|}{$\epsilon = 0.05$} & \multicolumn{2}{c|}{$\epsilon = 0.1$} & \multicolumn{2}{c|}{$\epsilon = 0.2$} & \multicolumn{2}{c}{$\epsilon = 0.3$} \\
 & ($\tau$) & Accuracy & Avg. $||\delta||_2$ & Accuracy & Avg. $||\delta||_2$ & Accuracy & Avg. $||\delta||_2$ & Accuracy & Avg. $||\delta||_2$ \\
    \hline
    \multirow{3}{*}{METEOR} & 0.15 & 74.6 & 2.6162 & 82.3 & 3.8674 & 88.4 & 5.4810 & 95.3 & 10.6270 \\
    & 0.20 & 69.5 & 2.8773 & 75.3 & 4.1655 & 79.8 & 5.5615 & 86.7 & 10.9520 \\
    & 0.25 & 58.6 & 2.9984 & 70.2 & 4.3788 & 74.3 & 5.7853 & 81.5 & 11.7785 \\
    \hline
    Exact Match & 1.0 & 39.3 & 4.5325 & 49.7 & 4.4385 & 57.2 & 5.8425 & 66.7 & 10.6545 \\
    \hline
    \end{tabular}%
  \label{tab:SAT}%
  \vspace{-3mm}
\end{table*}%

\subsection{Gray-Box Attack}
We use gray box setup in our experiments where we do not have any details of the language model of the captioning framework. 
We perform experiments with known CNN \ie,~feature extractor but no information regarding 
model's augmentation techniques \eg,~attention or semantic attributes. This is in sharp contrast to the existing methods that take advantage of attacked model's objective function and perform the attack~\cite{chen2017attacking, xu2018fooling, xu2019exact}.  

\vspace{1mm}
\noindent \textbf{Targeted Caption Results:}
In this section we evaluate our attack on the  popular
state-of-the-art  image captioning models \ie,~Show and Tell~\cite{vinyals2015show}, and Show-Attend-and-Tell~\cite{xu2015show}. The models vary in their internal architecture. For instance, Show, Attend and Tell employs attention mechanism in their framework whereas Show and Tell does not use attention. 
To evaluate the attack, we use the MS COCO dataset~\cite{lin2014microsoft} which is used by the aforementioned  methods and vast majority of other existing methods. To the best of our knowledge, ours is the first and the only method that performs gray-box attack on the captioning framework. Hence, the reported results do not include existing method (there are none in this direction). All the existing methods on image captioning use language model information and devise their attack strategy based on the adversary's objective functions. In contrast, we manipulate the image caption by an attack completely based on the visual model of the captioning framework. 

To evaluate the attack, first we select all the images from the source class and randomly select one image from the target class. Note that, since existing methods employ the adversary's language model objective function, they must use the target image's caption. In contrast, we do not have any information regarding the adversary's language model. Therefore, we do not use target image's caption as a target.  Instead, we directly use the target image itself for perturbation generation, and fool the framework to generate the caption of the source image similar to that of the target image. 

For the generated adversarial captions, we evaluate the attack success rate by measuring  all the predictions of the class dataset that match the target captions. Note that,  unlike image classification where the image label is predefined, the correct or relevant caption for an image has unlimited space. The two captions for an image may be altogether different yet correct. Therefore, we believe that exact match is an inappropriate  criterion which may treat a semantically correct caption as incorrect. We resolve this issue by employing METEOR score as the metric for match between the two predictions. METEOR has proven to be robust metric which not only caters for all the possible synonyms of the words but it is found to be  highly correlated with human judgments in settings with low number of reference~\cite{vedantam2015cider}. As described below, we devise two metrics for caption matching.

\noindent 1. {\tt Exact-Match:} The two predictions \ie,~of source image and target image are identical. \\
2. {\tt METEOR Score} $> \tau$: The metric measures whether the METEOR score between the captions exceeds beyond certain threshold. Inspired by evaluation metrics in dense image/video captioning~\cite{johnson2016densecap, krishna2017dense} we use METEOR score thresholds 
$0.15, 0.20, 0.25$. 

\vspace{1mm}
\noindent\textbf{Attack results on Show and Tell Model}~\cite{vinyals2015show} are shown in Table~\ref{tab:SAAT}. From the experiments, we note that the accuracy of the derived metric from METEOR is higher than that of the Exact Match which is intuitive. We see that for the exact match the accuracy varies from $39.3\%$ to $66.7\%$ for different $\epsilon$ values. Considering the gray-box constraint and having no information at all regarding the language model,  the success rate is significant. As the exact match is a highly constrained metric for captions comparison (as  two captions may be altogether different but correct at the same time), we see that derived METEOR metric success rate varies from $81.5\%$ up to $95.3\%$.

\begin{figure*}[t]
  \centering
     \includegraphics[width=0.75\textwidth]{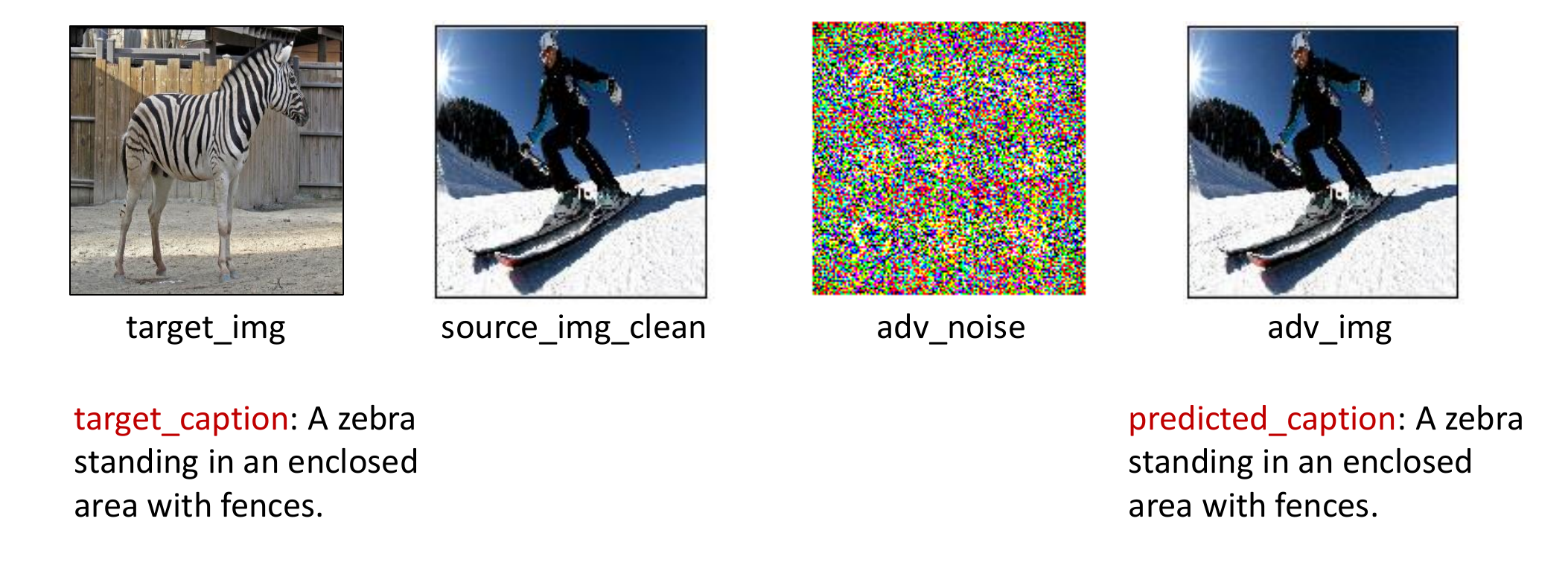}
     \vspace{-3mm}
     \caption{A qualitative example of the proposed adversarial attack on the Show-Attend-and-Tell~\cite{xu2015show} model. The example shows successful Exact attack where the model is fooled to generate the exact caption of the target image given the source image (on which the attack is launched).}
      \label{fig:qual_fig_1}
      \vspace{-2mm}
\end{figure*}

\vspace{1mm}
\noindent \textbf{Attack results on Show-Attend-and-Tell Model}~\cite{xu2015show} are presented in Table~\ref{tab:SAT}. We see from the results that attack performance is slightly better as compared to Show-and-Tell. The possible reason is that the model structures of Show-and-Tell and Show-Attend-and-Tell are significantly different from each other. Specifically, in Show-and-Tell model, the extracted visual features by the CNN are fed at the starting time step only. However, Show-Attend-and-Tell model feeds the extracted CNN features at every time step and employs attention mechanism in its language model. This means the model heavily relies on the visual features input. Thus, the   perturbations added to the visual features result in better attack performance. Note that, this observation is somewhat different than what  is reported in literature~\cite{xu2019exact} \ie,~ the attack performance  significantly worsens on Show-and-Tell model \ie,~from $99.56\%$ to $44.04\%$. We believe that the observed phenomenon  is due to the variance in the internal mechanism of the two models. The technique~\cite{xu2019exact} back-propagates the gradients from the observed words to the input image, which can be directly done in the case of Show-Attend-and-Tell. However, in the case of Show-and-Tell, they need to multiply the latent and observed words gradients first and then back-propagate to the input image. This, in turn, results in the poor performance of the attack on relatively simpler model. However, we do not suffer from this bottleneck as we do not use language model objective function in our technique. For the qualitative example please see Fig,~\ref{fig:qual_fig_1}.


\begin{table}[t]
  \small
  \centering
  \setlength{\tabcolsep}{4.0 pt}       
  \caption{Percentage of partial success with different $\epsilon$ values in the failed attack examples. Average of $||\delta||_2$ is computed over all the examples falling in each category.}
    \begin{tabular}{c|c|c|c|c}
    \hline
    \multirow{2}[4]{*}{} & \multicolumn{3}{c|}{Accuracy} & \multirow{2}{*}{Avg. $||\delta||_2$} \\
          & 1-keyword & 2-keyword & 3-keyword &  \\
    \hline
    \hline
    $\epsilon=0.05$ & 93.7\% & 90.1\%  & 88.4\%  & 1.8594 \\
    \hline
    $\epsilon=0.1$ & 94.2\% & 91.6\% & 89.3\% &  2.6397\\
    \hline
    $\epsilon=0.2$ &  94.9\%  & 92.4\% & 90.1\% &  4.9437\\
    \hline
    $\epsilon=0.25$ & 96.2\% & 92.5\% & 90.7\% &  6.7724\\
    \hline
    \end{tabular}%
  \label{tab:keyword}%
\end{table}%

\begin{figure}[t]
  \centering
     \includegraphics[width=0.8\columnwidth]{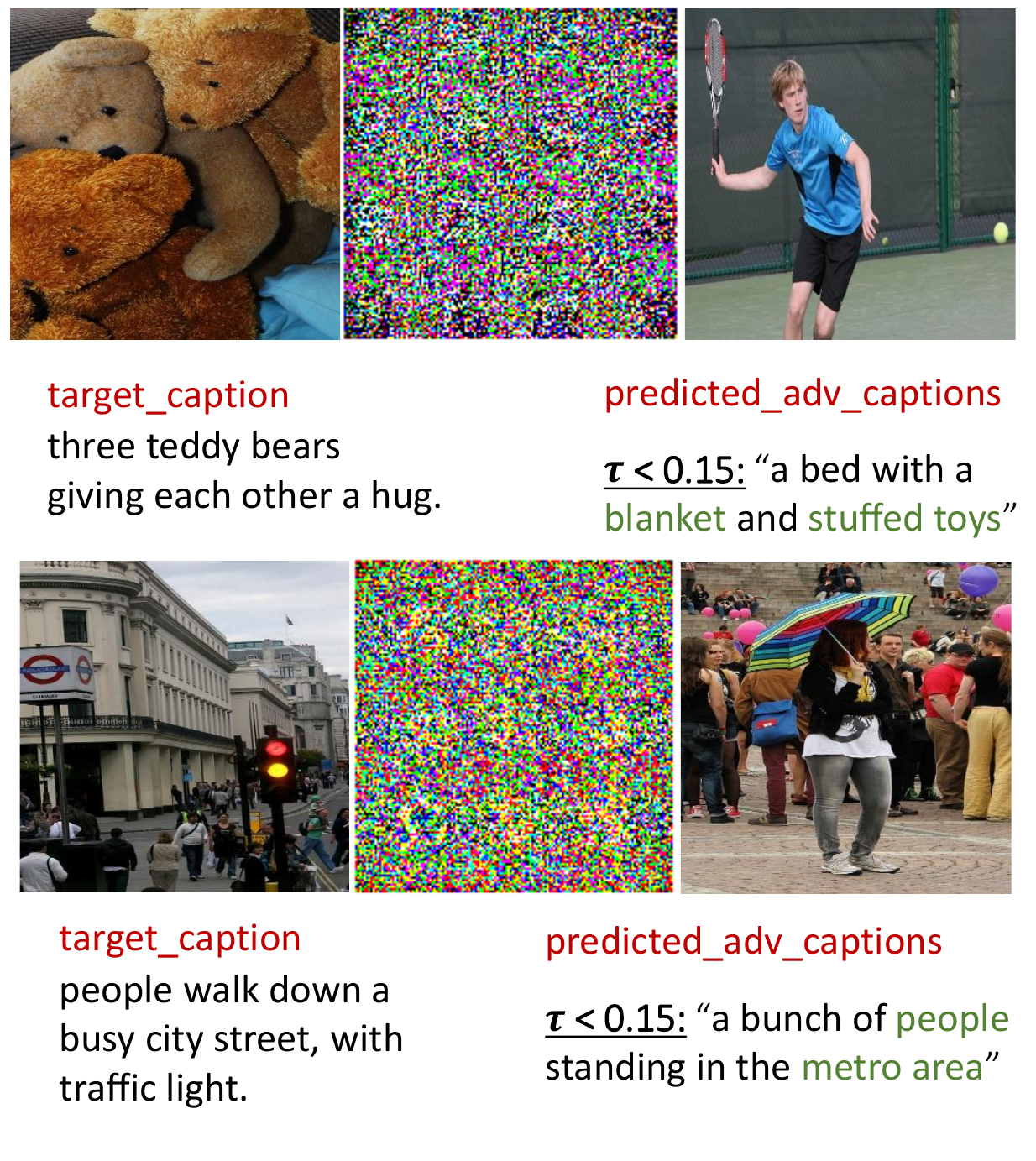}
     \vspace{-3mm}
     \caption{Examples of not (fully) successful attacks with target image and selected caption (left), and source image with predicted caption (right). The predicted captions fall under the minimum threshold criteria of $\tau$ = 0.15, and hence are  considered unsuccessful. However, it can be observed that though the predicted captions have low metric score, they are still relevant and semantically meaningful. }
      \label{fig:qual_res_failed_2}
      \vspace{-3mm}
\end{figure}

\begin{table}[t]
  \small
  \centering
  \setlength{\tabcolsep}{8.0 pt}       
  \renewcommand{\arraystretch}{1.35}  
  \caption{Performance comparison of our technique with the only available method Show-and-Fool~\cite{chen2017attacking} reporting model performance by attacking  the visual encoder. The scores reported from [7] are based on attacks on classification layer in comparison to our technique where we attack the internal layer of the visual encoder. Both the methods employ Show-and-Tell~\cite{vinyals2015show} model.}
    \begin{tabular}{c|c|c}
    \hline
    Experiment & \multicolumn{1}{c|}{Accuracy} & \multicolumn{1}{c}{Avg. $||\delta||_2$ } \\
    \hline
    \hline
    Show-and-Fool~\cite{chen2017attacking} [C\&W] & 22.4\% & 2.870 \\
    \hline
    Show-and-Fool~\cite{chen2017attacking} [I-FGSM] & 34.5\% & 15.596 \\
    \hline
    Ours  & \textbf{66.7\%} & 6.555 \\
    \hline
    \end{tabular}%
  \label{tab:cnnonly}%
  \vspace{-5mm}
\end{table}%
\vspace{1mm}
\noindent \textbf{Failed Attacks and Target Keyword:}
In addition to the complete caption results, as stated above, we see the failed attacks and observe whether the predictions include the keywords in it or not. We select the intersection of the top-3 predictions by the image classifier and the target caption as the keywords. We consider the attack still successful if the predicted captions include the target keyword(s). The results are reported in Table~\ref{tab:keyword}. As evident from the table, almost $94\%$ of the failed examples include at least one keyword in the predicted caption and almost $90\%$ of the examples contain $3$ keywords. Moreover, though the predicted captions achieve low score on automated metric but we  see from the examples that the predicted captions are still relevant to target image content and description (see Fig.~\ref{fig:qual_res_failed_2} for qualitative examples). These results depict that even failed attempts are still acceptably successful. 

\begin{table*}[t]
  \small
  \setlength{\tabcolsep}{6.0 pt}       
  \centering
  \caption{Comparisons of adversarial attacks with $\epsilon$ = [0.05, 0.1, 0.2, 0.25] on two vanilla image captioning frameworks with hidden state-size (SS) of 512 (left) and 2048 (right) on validation set of MS COCO dataset~\cite{lin2014microsoft}. B-[1-4], M, C, R, and S represent evaluation metrics BLEU-[1-4], METEOR, CIDEr, ROUGE and SPICE, respectively.}
  \vspace{1mm}
    \begin{tabular}{c|cccccccc||cccccccc}
    \hline
    \multirow{2}{*}{Val Set} & \multicolumn{8}{c||}{Baseline-Model-I (SS-512)}  & \multicolumn{8}{c}{Baseline-Model-II (SS-2048)} \\
    \multicolumn{1}{c|}{} & B1 & B2 & B3 & B4 & M & C & R & S & B1 & B2 & B3 & B4 & M & C & R & S \\
    \hline
    \hline
    clean-set & 65.3 & 46.9 & 32.8 & 22.9 & 22.3 & 77.0 & 48.0 & 15.5 & 63.1 & 45.5 & 29.2 & 19.6 & 20.8 & 67.9 & 45.7 & 14.4 \\
    \hline
    \hline
    $\epsilon = 0.05$   & 48.7 & 28.2 & 16.3 & 9.9 & 13.7 & 23.6 & 36.2 & 6.3 & 46.2 & 25.0 & 13.5 & 7.6 & 12.6 & 18.8 & 34.3 & 5.2 \\
    \hline
    $\epsilon = 0.10$   & 46.2 & 25.4 & 13.8 & 8.0 & 12.4 & 16.8 & 34.5 & 4.8 & 44.3 & 23.1 & 11.8 & 6.4 & 11.7 & 13.9 & 33.1 & 4.0 \\
    \hline
    $\epsilon = 0.20$   & 44.1 & 23.1 & 11.9 & 6.5 & 11.4 & 12.5 & 33.0 & 3.8 & 42.6 & 21.0 & 10.2 & 5.3 & 10.8 & 10.7 & 31.9 & 3.2 \\
    \hline
    $\epsilon = 0.30$   & 42.9 & 21.9 & 10.9 & 5.8 & 10.9 & 10.1 & 32.2 & 3.3 & 41.8 & 20.3 & 9.8 & 5.1 & 10.3 & 8.6 & 31.4 & 2.9 \\
    \hline
    \end{tabular}%
  \label{tab:statesize}%
  \vspace{-3mm}
\end{table*}%
\subsection{CNN Only Attack}
As our technique attacks on the CNN part of the image captioning framework, we compare our method with the only known method Show and Fool~\cite{chen2017attacking} reporting results of attacks on CNN-only, employing the popular I-FGSM~\cite{kurakin2016adversarial} and C\&W~\cite{carlini2017adversarial} attacks. In Table~\ref{tab:cnnonly} we present the comparison between Show-and-Fool and our method on the attack on CNN only. Show-and-Fool employed Tensorflow implementation of the Show-and-Tell~\footnote{\url{https://github.com/huanzhang12/ImageCaptioningAttack}} model with Inception-V3 as CNN part. Moreover, it carefully selects $800$ images in such a way that it has at least one word common with Show-and-Tell vocabulary with $100\%$ classification accuracy on Inception-v3. In contrast, we do not set any of these constraints. For a fair comparison, we adopt the same CNN model \ie,~Inception-v3 in our implementation and experimental hyperparameter values such as $\epsilon$. Despite that  Show-and-Fool sets very weak criterion of success, we see that our method outperforms it with a significant margin. The results confirm that attacking the image captioning framework (\ie,~CNN+RNN) is inherently more challenging than attacking the image classifier, and it requires more careful designing of the adversarial examples. As seen from the results, attacking the image class does not transform well in image captioning framework. However, devising the attack based on the internal layers of the classifier results in much better performance.

\subsection{Internal Layer Size Effect on Attack Success}
In this section, we investigate the question that does the number of units in an LSTM network affect the attack performance? To explore this, we design two vanilla image captioning frameworks without any sophistication such as attention or semantic attributes \etc. We keep all the parameters and training settings similar for both the models. Except one parameter \ie,~hidden layer size of the LSTM. We select two sizes $512$ and $2048$ for the two models in our experiments. Moreover, we perform four experiments on each model using  $\epsilon = \{0.05, 0.1, 0.2, 0.25\}$. The predicted captions for the whole validation dataset are then evaluated using the popular automated evaluation metrics \ie,~BLEU-[1-4], METEOR, CIDEr, ROUGE and SPICE. The results of this experiment are reported in Table~\ref{tab:statesize}. For upper bounds of the model, we first compute the metric scores using the benign image set as shown in the first row of the Table~\ref{tab:statesize}. We then run the model with adversarial images perturbed with various $\epsilon$ values and compute the metric scores for all predictions. Interestingly, the model with hidden state size of $512$ performs slightly better on all runs. However, we see that performance of both the models degrade with increase in the $\epsilon$ value. The trend is similar in both the variants of the model. We analyse another aspect from the experiment  regarding the behaviour of the evaluation metrics. We see that the B-1 that refers to {\em uni-grams} has little effect for all four attacks. It means that overall words in the caption do not change much. The same effect is corroborated in our keyword experiments where we observe  that even from the failed attacks almost $94\%$ of the captions contain the keywords.

From  Table~\ref{tab:statesize}, we see that both the models show similar trend against the adversarial attack. Thus, we can conclude that the internal state size does not play a major role as a defense against the adversarial attack. The slight difference in both model's performance is inherent due to the state size mismatch, as evident from the results in the first row computed on benign images.     


\vspace{-2mm}
\section{Conclusion}
\vspace{-1mm}
In this work, we studied the adversarial attack on more complex image captioning framework (\ie,~CNN+RNN) and proposed an algorithm to generate adversarial examples for the framework. The algorithm performs gray-box type attack as it does not require the targeted captioning framework's language model or other internal details. The proposed algorithm is a first-of-its-kind that controls the predicted captions by attacking only on the visual encoder of the captioning framework. Our experiments show that the proposed method can generate  adversarial examples against any state-of-the-art model with a high attack success rate. We also demonstrate that attacking the image captioning system is  different from attacking image classifiers. Moreover, attack on the classification layer of the feature extractor results in poor success rate in comparison to attacking the internal layer of the feature extractor. Further we observed that the internal state size of the recurrent network does not play a major role in robustification of the model against the adversarial attacks. The proposed method stands different from the contemporary methods in the sense that we control the captions by attacking only the visual encoder. To the best of our knowledge, this is the very first work on crafting adversarial examples under such a gray-box setting for image captioning systems. 

{\small
\bibliographystyle{ieee_fullname}
\bibliography{egbib}
}

\end{document}